\documentclass[10pt,twocolumn,letterpaper]{article}
\usepackage[pagebackref,breaklinks,colorlinks]{hyperref}

\usepackage[pagenumbers]{cvpr} % To force page numbers, e.g. for an arXiv versio

\usepackage{graphicx}
\usepackage{amsmath}
\usepackage{amssymb}
\usepackage{amsfonts}
\usepackage{booktabs}
\usepackage{graphicx}
\usepackage[normalem]{ulem}
\usepackage{amsmath, amssymb}
\usepackage{footnotebackref}
\usepackage[para]{footmisc}
\usepackage{color}
\usepackage{enumerate}
\usepackage{graphbox}
\usepackage{makecell}
\usepackage{soul}
\usepackage{wrapfig}
\usepackage{graphics}
\usepackage{subcaption}
\usepackage{multirow}
\usepackage{eqparbox,array}
\usepackage{enumerate}
\usepackage{booktabs}
\usepackage{isotope}
\usepackage{siunitx}
\usepackage{tablists}
\usepackage{epsfig}
\usepackage{bm}
\usepackage[table]{ xcolor}
\usepackage{enumerate}
\usepackage{authblk}
\usepackage{enumitem}
\usepackage{cite}

\usepackage{amsmath}
\usepackage{amssymb}
\usepackage{amsfonts}
\usepackage{textcomp}
\usepackage{algorithm}
\usepackage{enumitem}
\usepackage[noend]{algpseudocode}
\usepackage{threeparttable}
\usepackage{booktabs}       % professional-quality tables
\usepackage{amsfonts}       % blackboard math symbols
\usepackage{nicefrac}       % compact symbols for 1/2, etc.
\usepackage{microtype}      % microtypography
\usepackage{subcaption}
\usepackage{cleveref}

\usepackage[pagebackref,breaklinks,colorlinks]{hyperref}
\makeatletter
\renewcommand\AB@affilsepx{, \protect\Affilfont}
\makeatother
\usepackage{lipsum}
\newcommand\blfootnote[1]{%
\begingroup
\renewcommand\thefootnote{}\footnote{#1}%
\addtocounter{footnote}{-1}%
\endgroup
}

\newcommand{\red}[1]{\textcolor{black}{#1}}

\usepackage{indentfirst}
\usepackage[authoryear,longnamesfirst]{natbib}
\setcitestyle{numbers,square}

\def\cvprPaperID{6786} % *** Enter the CVPR Paper ID here
\def\confName{CVPR}
\def\confYear{2022}

\definecolor{darkgreen}{rgb}{0,0.6,0.2}
\definecolor{danred}{rgb}{0.9098,0.9098,0.9098}
\definecolor{shenred}{rgb}{0.8117,0.8117,0.8117}

\newcommand{\RN}[1]{\textup{\lowercase\expandafter{\it \romannumeral#1}}}

\begin{document}

%%%%%%%%% TITLE - PLEASE UPDATECCbeathumanbeathumanbeathuman
\title{elBERto: Self-supervised Commonsense Learning for Question Answering} %Inter-Modality
\author[1$\dagger$]{Xunlin Zhan}
\author[2$\dagger$]{Yuan Li}
\author[3]{Xiao Dong}
\author[1$\star$]{Xiaodan Liang}
\author[4]{Zhiting Hu}
\author[2]{Lawrence Carin}

\affil[1]{Shenzhen Campus of Sun Yat-sen University}
\affil[2]{Duke University}
\affil[3]{Sun Yat-sen University}
\affil[4]{University of California
\protect\\
\textit{\small  \{zhanxlin,dongx55\}@mail2.sysu.edu.cn, xdliang328@gmail.com, zhitinghu@gmail.com,
\{yl558,lcarin\}@duke.edu }}

\maketitle

%%%%%%%%% ABSTRACT
\begin{abstract}
Commonsense question answering requires reasoning about everyday situations and causes and effects implicit in context.
% Typically, current systems use large language models (LMs) pre-trained with self-supervised learning (SSL), and fine-tune in a supervised manner on downstream QA datasets.
Typically, existing approaches first retrieve external evidence and then perform commonsense reasoning using these evidence.
% Despite the improved results, the systems still fall short of accurate inference of commonsense knowledge underlying the text due to the intrinsic difference between commonsense context and general text.
In this paper, we propose a
S\textbf{el}f-supervised \textbf{B}idirectional \textbf{E}ncoder \textbf{R}epresen\textbf{t}ation Learning of C\textbf{o}mmonsense (elBERto)
% pipeline
framework, which is compatible with off-the-shelf QA model architectures. The framework comprises five self-supervised tasks to force the model to fully exploit the additional training signals from contexts containing rich commonsense.
The tasks include a novel Contrastive Relation Learning task to encourage the model to distinguish between logically contrastive contexts, a new Jigsaw Puzzle task that requires the model to infer logical chains in long contexts,and three classic self-supervised learning(SSL) tasks to maintain pre-trained models' language encoding ability.
On the representative WIQA, CosmosQA, and ReClor datasets, elBERto outperforms all other methods using the same backbones and the same training set, including those utilizing explicit graph reasoning and external knowledge retrieval. Moreover, elBERto achieves substantial improvements on out-of-paragraph and no-effect questions where simple lexical similarity comparison does not help, indicating that it successfully learns commonsense and is able to leverage it when given dynamic context.

\blfootnote{$\dagger$ Equal contribution. $\star$ Corresponding Author.}
\end{abstract} %\textbf{I}nter-\textbf{M}odality

%%%%%%%%% BODY TEXT
\section{Introduction}

\label{intro}

\begin{figure*}[!htb]
    \centering
    \includegraphics[trim=0 210 0 0, clip, scale=0.69]{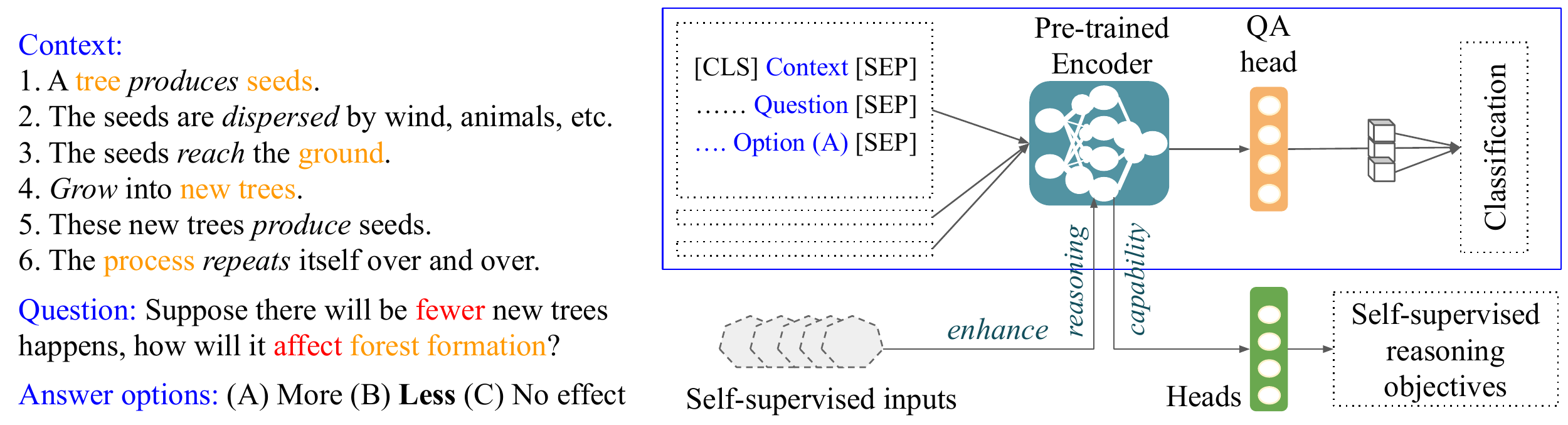}
    \caption{
    {\bf Left:} An example of commonsense multiple-choice QA. The context contains complex relations among entities (orange text) and dynamically changing processes (italic words). The question asks about the logical influences of changing events or processes.
    {\bf Right:}
    Illustration of the proposed elBERto pipeline, where the modules in the blue box form the model backbone, and the rest demonstrate the components for self-supervised commonsense learning. All objectives share the same language encoder to encourage explicitly capturing commonsense knowledge.
    }
    \label{fig:example}
    % \vspace{-2mm}
\end{figure*}

The commonsense learning is a key aspect in learning general knowledge.
The commonsense question answering (QA) task aims to examine QA systems' ability to engage in commonsense reasoning in a context that contains complicated logical relationships and implicit universal knowledge.
A typical commonsense multiple-choice QA example contains a context, a question, and multiple answer options---among which one and only one is correct---as demonstrated in Figure~\ref{fig:example} (left).
The context provides background support for reasoning and has been proposed in various complex forms, such as procedural descriptions about the change of events and processes~\cite{tandon2019wiqa}, people's everyday narratives containing causes and effects~\cite{huang2019cosmos}, and reading comprehension problems in standardized tests comprising logical rules and potential conflicts~\cite{yu2020reclor}.
In this setting, a commonsense QA solver aims to choose the most plausible correct answer from multiple options based on the context and question. However, it is challenging to solve commonsense QA because questions are intentionally created to examine out-of-context knowledge, and simple similarity matching between questions and the context does not work.

Recently, pre-trained language models, such as BERT~\cite{devlin2018bert}, RoBERTa~\cite{liu2019roberta}, and XLNet~\cite{yang2019xlnet} have achieved new state-of-the-arts in many natural language processing (NLP) tasks~\cite{huang2019cosmos,tandon2019wiqa,zhu2020incorporating}. These models learn generic language representation through self-supervised learning (SSL) on a massive scale of text corpora.
Unsurprisingly, on the commonsense QA task, pre-trained models outperform many well-established neural models,
even without context or when tested on unseen topics~\cite{tandon2019wiqa}.
% This suggests that these models learn some form of implicit commonsense knowledge.
However, most existing SSL tasks do not explicitly model commonsense knowledge, where the context has intrinsic differences from general text. Thus, a large gap still remains between pre-trained models' performance and human benchmarks.
% indicating significant room for improving logical reasoning over context that contains complicated multi-hop causes and effects.
Several attempts have been made to alleviate this issue, such as adapting Transformers to encode extra hops of a textual evidence graph in Transformer-XH~\cite{zhao2020transformer-xh}, injecting knowledge through an attention mechanism~\cite{ma2019towards}, and traversing a virtual knowledge base~\cite{dhingra2020differentiable}. However, the first attempt suffers from learning implicit rules that may contain data bias, whereas the second and third attempts require a carefully selected knowledge base that contains substantial overlap with the dataset domain and potentially drastically elongated (pretext) training.

To investigate the extent to which SSL can be leveraged for explicit commonsense modeling,
we propose a  S\textbf{el}f-supervised \textbf{B}idirectional \textbf{E}ncoder \textbf{R}epresen\textbf{t}ation Learning of C\textbf{o}mmonsense (elBERto) pipeline that facilitates explicit commonsense learning and reasoning through Transformer's bidirectional encoder architecture. elBERto is equipped with two novel self-supervised tasks and three complementary tasks.
First, we propose a novel Contrastive Relation Learning (CRL) task that requires a QA solver to distinguish between a logically valid context and an invalid one with edited but similar content by alternating significant words in the context to their antonyms. Correctly identifying the semantically valid context requires the QA solver to fully understand the local influences between events and commonsense logic. Second, we propose a Jigsaw Puzzle (JP) task to encourage a better understanding of logical chains in long contexts; specifically, the QA solver is presented with multiple sentences extracted from a context and randomly shuffled and optimized toward predicting the correct permutation. Further, we incorporate three SSL tasks proven to be effective in previous works~\cite{lan2019albert,jernite2017discourse,zhang-etal-2019-ernie,joshi2020spanbert} for better sentence and entity understanding, including binary sequence order prediction (BSOP), masked language modeling (MLM), and masked entity modeling (MEM).

elBERto has three advantages, which are delineated as follows:
\begin{itemize}
    \item it utilizes the in-domain training corpus in a data augmentation manner without external data, providing more training signals and achieving data efficiency;
    \item it does not require extra pretext training time compared to existing methods; and
    \item it reduces the domain gap between seen and unseen context~\citep{sun2019unsupervised,sun2019test} by learning universal commonsense, improving transferability and interpretability.
\end{itemize}

We conduct comprehensive experiments on elBERto in terms of the three following challenging commonsense QA tasks: WIQA~\cite{tandon2019wiqa}, CosmosQA~\cite{huang2019cosmos} and ReClor~\cite{yu2020reclor}. Qualitative and quantitative experiments demonstrate that each proposed SSL task leads to performance improvement over baseline, and elBERto effectively fuses the five tasks, achieving the best accuracy in all datasets. Further, CRL and JP bring substantial improvement, especially on relatively difficult questions such as out-of-paragraph questions of WIQA and hard-level questions of ReClor. Finally, on both the CosmosQA and ReClor leaderboards, elBERto exceeds all other methods using the same backbones and the same training set, including those utilizing explicit graph reasoning and external knowledge retrieval.

The main contributions of this paper can be summarized as follows:
\begin{itemize}
\item We make the first attempt to present five SSL tasks to empower pre-trained language models with a strong capacity to understand rich logics and knowledge for commonsense QA.
\item We propose a novel CRL task and a JP task to facilitate explicit
commonsense learning and reasoning. 
% \item \FIXME{We propose a novel CRL task to help QA solvers to identify the valid context and a JP task to choose the correct order of even permutations.}
\item Our simple yet effective pipeline surpasses all existing works using the same backbones and the same training set in the three most challenging commonsense QA benchmarks, even those with explicit graph reasoning and knowledge retrieval.
\end{itemize}

The rest of the paper is organized as follows: In Section ~\ref{sec: related}, we give a detailed synopsis of related work, including commonsense reasoning, pre-trained language models, and SSL. Sections ~\ref{sec: problem} and ~\ref{sec: method} introduce the core idea of our proposed model and the corresponding optimization objectives, respectively. The experimental results are discussed in Section ~\ref{sec: exper}. Finally, Section ~\ref{sec: conclusion} concludes and presents our future work.

\section{Related Work}
\label{sec: related}

% Our elBERto draws inspiration both from the aspect of pre-trained language models and self-supervised learning to solve commonsense question answering task. In what follows, we briefly review existing studies on Commonsense QA, pre-trained language models and learning with auxiliary tasks.

\subsection{Commonsense QA}
Commonsense QA~\cite{huang2019cosmos,tandon2019wiqa,yu2020reclor} requires understanding and reasoning over real-world knowledge.
Most previous approaches to commonsense QA fall into the two following categories:
\begin{itemize}
    % \item \red{Retrieving corresponding evidence or paths from an external knowledge base, serving as data augmentation~\citep{huang2020rem, ZHAN-pathreasoner, wang-etal-2021-k-adapter}. For example,  \citep{huang2020rem} proposed a recursive erasure memory network termed REM-Net to cope with the quality improvement of evidence. K-adapter~\citep{wang-etal-2021-k-adapter} infuses knowledge into pre-trained models using adapters. GDIN~\citep{tian-etal-2020-scene-gdin} utilizes a external knowledge base to restore scene during reading narrative for better comprehension. RekNet~\citep{zhao2022reference-RekNet} refines critical information from the passage and quote explicit knowledge in necessity.}
    \item Retrieving corresponding evidence or paths from an external knowledge base, serving as data augmentation~\citep{huang2020rem, ZHAN-pathreasoner, wang-etal-2021-k-adapter}. For example,  \citep{huang2020rem} proposed a recursive erasure memory network termed REM-Net to improve the evidence quality. \citep{wang-etal-2021-k-adapter} infuses knowledge into pre-trained models using a specific adapter, K-adapter. GDIN~\citep{tian-etal-2020-scene-gdin} utilizes an external knowledge base to restore scenes during reading narrative for better comprehension. RekNet~\citep{zhao2022reference-RekNet} refines critical information from the passage and quotes explicit knowledge in necessity.
    \item Establishing context-based graphs with nodes as the keywords extracted from context, then propagating the information between nodes via Graph Neural Networks (GNNs)~\citep{kipf2016semi,kundu2018exploiting,mihaylov2018knowledgeable,tu2019multi,huang2021dagn}. For example, CURIE~\citep{Rajagopal2021CURIEAI} iteratively constructs an explicit graph of relevant consequences for commonsense reasoning in a structured situational graph utilizing natural language queries. RGN~\citep{ijcai2021-0553-RGN} proposes a relational gating network to learn to filter the key entities and relationships and learns contextual and cross representations of both procedure and question for finding the answer.
\end{itemize}
There are also approaches falling into both categories: In KagNet~\citep{lin2019kagnet}'s study, the system retrieves external evidence and then performs commonsense reasoning via GNNs.
% There are also approaches falling in both categories~\citep{lin2019kagnet} which retrievals external evidence and then perform commonsense reasoning via GNNs.
In addition, there are some methods using data argumentation or regularized optimization. Logic-Guid~\citep{asai2020logic} is presented to enrich the training set by replacing words with their antonyms and leveraging logical and linguistic knowledge to augment labeled training data. ALICE~\citep{pereira-etal-2020-adversarial} uses an adversarial training algorithm to achieve commonsense inference. SMART~\citep{jiang-etal-2020-smart} uses principled regularized optimization to perform robust and efficient fine-tuning for pretrained models. However, these works require time-consuming retrieval of a large amount of evidence, which is prone to containing noise. Further, GNNs only propagate the information between nodes that have associated effects but fail to involve intervention and counterfactual reasoning. We propose a lightweight pipeline that is capable of performing intervention and counterfactual reasoning, achieving better performance than GNNs.

\subsection{Pre-trained Language Models (LMs)}
The aim of pre-training LMs is to learn universal language representations via web-scale unlabeled data and pretext tasks. Recently, significant pre-trained models, such as BERT~\citep{devlin2018bert}, RoBERTo ~\citep{liu2019roberta}, XLNet~\citep{yang2019xlnet}, ALBERT~\citep{lan2019albert} and ERNIE~\citep{li2019enhancing} have been shown to be effective for downstream NLP tasks.
BERT~\citep{devlin2018bert} randomly masks out some of the words and trains the model to predict the masked tokens and whether the next sentence will be coherent.
In ~\citep{lan2019albert}'s work, two parameter-reduction techniques are presented to lower memory consumption and increase the training speed of BERT.
~\citep{li2019enhancing} propose a word-aligned attention to exploit explicit word information.
However, this is not trained specifically for commonsense learning and reasoning over long, complicated contexts.
Some recent attempts to adapt pre-trained LMs to commonsense QA ({\em e.g.},  Zhang $\emph{et al.}$~\citep{zhang-etal-2019-ernie} and Xiong $\emph{et al.}$~\citep{xiong2019pretrained}) propose to incorporate entity-level knowledge, while Sun $\emph{et al.}$~\citep{Sun2019ERNIEER} utilize knowledge graphs.
Lan $\emph{et al.}$~\citep{lan2019albert} used a self-supervised loss that focuses on modeling inter-sentence coherence.
However, these approaches require substantial training data and extensive pretext training time. Further, they rely heavily on external knowledge bases. In contrast, elBERto requires little training overhead by leveraging the existing training corpus.

\subsection{Learning with Auxiliary Tasks}
Self-supervised auxiliary tasks have been widely adopted in the field of machine learning.
~\citep{Zhu0CL20} introduce an auxiliary reasoning navigation framework with four self-supervised auxiliary reasoning tasks to take advantage of the additional training signals derived from the semantic information.
In reinforcement learning, self-supervised auxiliary tasks are applied to improve data efficiency and robustness~\citep{veeriah2019discovery,ma2019self,jaderberg2016reinforcement}.
\citep{ma2019self} introduce a self-monitoring agent with a visual-textual co-grounding module to locate the instruction completed in the past and a progress monitor to ensure the grounded instruction. ~\citep{veeriah2019discovery} present a novel method for a reinforcement learning agent to discover questions formulated as general value functions.
In the NLP domain, SSL has been widely adopted as auxiliary tasks to pre-train a language model~\citep{devlin2018bert,lan2019albert}.
Logeswaran $\emph{et al.}$~\citep{logeswaran2018sentence} present a model that learns the order of sentences in a paragraph, while
Jernite $\emph{et al.}$~\citep{jernite2017discourse} design three discourse-based objectives over unlabeled text to train a sentence encoder.
However, all these methods have problems going beyond representation learning to comprehensively understand entities' relationship, events' causal influence, and the ordering of all context sentences.

\section{Problem Formulation}
\label{sec: problem}
In this section, we introduce our setting of a commonsense multiple-choice QA task.
% However,  our method is generally applicable to other commonsense QA settings such as generative QA.
In commonsense multiple-choice QA tasks, an input usually contains a context, a question and multiple answer options, among which one and only one is correct, as demonstrated in Figure~\ref{fig:example} (left). We formulate the input context as $\textbf{c} = \{c_1, c_2, \dots, c_n\}$, the question as $\textbf{q} = \{q_1, q_2, \dots, q_m\}$, and a candidate answer as $\textbf{a} = \{a_1, a_2, \dots, a_k\}$, where $c_i$, $q_i$, $a_i$ represent the $i_{th}$ word of the context, question, and answer of lengths $n$, $m$, and $k$, respectively.
A QA system aims to choose the most plausible correct answer out of multiple answer options based on the context and the question. The objective can be written as

\begin{equation}
\small
\mathcal{L}_{QA} = - \frac{1}{N} \sum_{i=1}^{N} \log P(\textbf{a}_i^* | \textbf{c}_i, \textbf{q}_i, \{\textbf{a}_i^t \}_{t=1}^{T} \}),
\label{eq:qa}
\end{equation}

where $N$ is the number of training examples, $T$ is the number of answer options, and $\textbf{a}_i^*$ represents the ground truth answer of the $i_{th}$ example.

\section{Methods}
 \label{sec: method}
% \subsection{\FIXME{Overview} }

Figure~\ref{fig:example} (blue box at the upper right) illustrates our model framework.
% following prior works~\cite{zellers2018swag,talmor2018commonsenseqa,devlin2018bert}.
For each candidate answer $\textbf{a}_i^t$, the sequences of context, question, and $\textbf{a}_i^t$ are concatenated via a separator $\texttt{[SEP]}$. The whole sequence is prepended with a beginning-of-sequence token $\texttt{[CLS]}$ and appended with a end-of-sequence token $\texttt{[SEP]}$. An encoder module transforms each token in the input sequence to the summation of its corresponding word, position and type embedding, and then it encodes to a hidden representation. The hidden representation of the first token $\texttt{[CLS]}$ is then extracted to represent the semantic meaning of the whole sequence and further projected to a single logit via a linear module. The logits obtained by formatting and feeding all candidate answers into the model pipeline are then fed to a softmax layer and optimized by cross-entropy loss.
To complement the points that most pre-trained LMs are not explicitly developed for commonsense learning and that little training overhead is preferred, we propose leveraging five self-supervised tasks that approach the problem from distinctive angles. We explain each task below.

\subsection{Contrastive Relation Learning}

\begin{figure*}[!htb]
\center
\includegraphics[angle=0, width=1\textwidth]{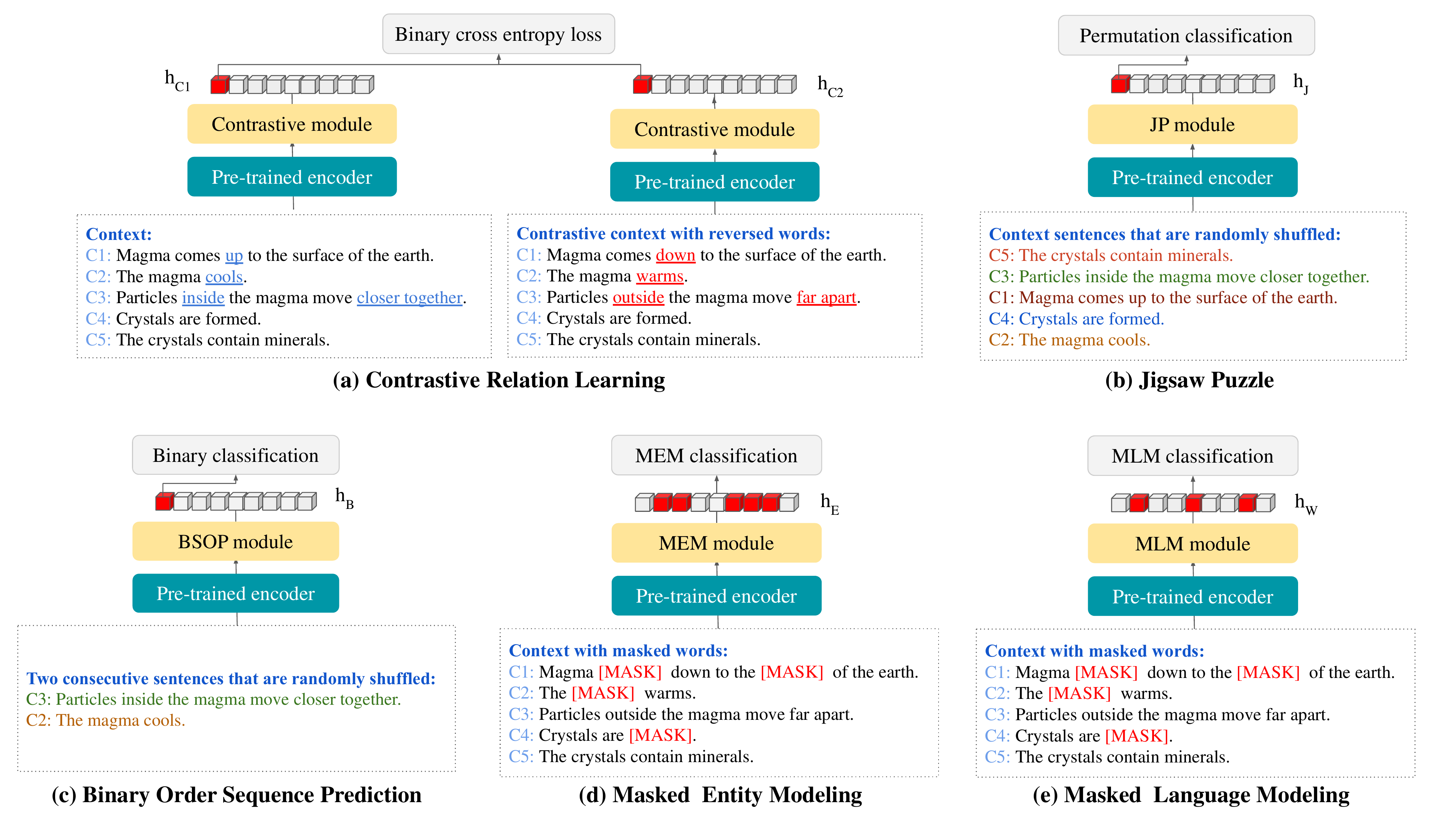}
\caption{Demonstration of the proposed Contrastive Relation Learning (CRL), Jigsaw Puzzle (JP), Binary Sequence Order Prediction (BSOP), Masked Entity Modeling (MEM), and Masked Language Modeling (MLM) SSL tasks.  All tasks share the same pre-trained encoder to encourage better capturing of commonsense knowledge.}
 \label{fig:self-supervised}
\end{figure*}

Commonsense QA contexts usually contain substantial logical influences between events. For example, in the context of Figure~\ref{fig:example}a, each sentence describes an event that is a logical consequence of the event described in the previous sentence ($\eg$ ``the seeds grow into new trees'' is caused by ``the seeds reach the ground''). Fully understanding such a context requires recognizing the causes and consequences between different events.
Moreover, it is observed that when changing one or multiple words in a context to their opposite meaning, the semantic meaning of each sentence may still be valid, but the local influences between sentences (events) are usually lost. For example, when replacing ``reach'' with its antonym ``move away,'' the resulting event---``the seeds move away from the ground''---is semantically reasonable, but it will not result in ``the seeds grow into new trees.'' While it is easy for humans to identify such implicit logical errors (seeds cannot grow without soil or nutrition), this is difficult for machines because there is no commonsense knowledge to refer to. To improve a QA solver's capacity to recognize commonsense logic, we design a CRL task by flipping significant words in a context to their antonyms; this generates contrastive context involving counterfactual logic.
Let $\textbf{c}_i$ and $\textbf{c}_i'$ denote the original and corresponding contrastive contexts, respectively. We train the QA solver to identify the valid context from $(\textbf{c}_i, \textbf{c}_i')$, which appear in random order; in this way, commonsense knowledge is learned. Further, we create a list of frequently occurring words associated with causes and effects for easy generation of pairs of contrastive contexts. Mathematically, the objective of CRL can be written as
\begin{equation}\small
\mathcal{L}_{CRL} = -\frac{1}{N}\sum_{i=1}^{N} \log P(\textbf{c}_i | (\textbf{c}_i, \textbf{c}'_i)).
\label{eq:crl}
\end{equation}
As illustrated in Figure~\ref{fig:self-supervised}a, the CRL task is implemented by appending a contrastive module to the pre-trained encoder for converting the encoded representation of each of a pair of contrastive contexts into one-dimensional logits. The logits are then fed to cross entropy loss for optimization. The encoder is shared between the CRL objective (Eq~\ref{eq:crl}) and the QA objective (Eq~\ref{eq:qa}) to empower the encoder with improved commonsense learning capability.

CRL utilizes a contrastive approach ({\em e.g.}, contrastive loss) that is close to contrastive learning of visual representation~\cite{chen2020simple,arora2019theoretical}, contrastive learning of graph embeddings~\cite{bordes2013translating,perozzi2014deepwalk,schlichtkrull2018modeling,velivckovic2018deep} and structured environment abstraction~\cite{kipf2019contrastive,franccois2019combined}.
However, elBERto differs from them in both the data formation and end goals. Previous approaches have constructed augmented samples and learned to identify these matched pairs from the other of contrastive samples to force the model to be invariant to data augmentation. elBERto constructs contrastive samples and trains the model to recognize the only logical sound one between them, aiming to force the model to capture commonsense logical rules.
In addition, while Oscar~\cite{li2020oscar} uses contrastive loss---similar to elBERto---to differentiate image representations from polluted representations, this model was initially proposed for object-semantics pre-training.
Finally, CRL possesses a prominent design merit for commonsense QA by forming contrastive contexts with reversed words because questions in QA tasks often ask about the change of effects by alternating words describing quantity or severity level. For example, the question in Figure~\ref{fig:example}a asks about the effects of ``fewer'' new trees, which contrasts with the event of ``more'' trees described in the context (seeds ``grow into new trees''). Thus, training a QA solver with the CRL objective facilitates the acquisition of commonsense knowledge that generalizes to unseen scenarios. 

\subsection{Jigsaw Puzzles}

Besides acquiring the commonsense knowledge through CRL, it is also important to understand indirect influences scattered across long contexts. For example, ``a tree produces seeds'' can lead to indirect results of ``more forest formation.'' However, without the capability of inferring the missing intermediate logical chains or reasoning over the provided context for the correct chain, the model cannot come to this conclusion.
However, the longer the influence chain is, the harder it is to trace the events and reasoning over effects~\cite{tandon2019wiqa}. To help alleviate this problem, we propose an JP task where segments in a paragraph are randomly permuted, and the model is trained to choose the correct order out of $K$ presented permutations. The segments are either one sentence or multiple contiguous short sentences. Correctly restoring the sentence order requires the model to understand not only the semantic meaning of each individual sentence but also the long-chain influences along with them, and thus, it must capture indirect implications in different lengths of sentences. Let $(\textbf{s}_1, \textbf{s}_2, \dots, \textbf{s}_m)$ represent a sequence of consecutive sentences and $(\textbf{s}^k_1, \textbf{s}^k_2, \dots, \textbf{s}^k_m)$ represent the $k_{th}$ permutation of the sequence, where $1 \le k \le K$. The objective of JP can be written as
\begin{equation}\small
\mathcal{L}_{JP} = -\frac{1}{N}\sum_{i=1}^{N} \log P(r | \{(\textbf{s}^k_1, \textbf{s}^k_2, \dots, \textbf{s}^k_m)\}_{k=1}^{K}),
\label{eq:jp}
\end{equation}
where $r$ is the index of the correct order. To incorporate the JP objective, as illustrated in Figure~\ref{fig:self-supervised}b, we append a JP module to the pre-trained encoder such that encoded features of each context with shuffled sentence order are fed to the module and converted to one-dimensional logits. The logits are then optimized via cross-entropy loss.

\subsection{Additional Tasks}

We incorporate three classic SSL tasks to facilitate understanding of causal relationships, entity relationships, and language semantics.

\subsubsection{Binary Sentence Order Prediction (BSOP)}
Predicting sentence order has been proposed for learning better sentence representation~\cite{lan2019albert} and discourse-level coherence~\cite{jernite2017discourse,li2014model,li2016neural}. Based on similar techniques, we adapt a BSOP task to commonsense QA for a better understanding of causal relationships. Since consecutive sentences in QA examples usually exhibit strong causal relationships, training a QA solver to identify the correct order between two continuous sentences can lead to a better understanding of context. To achieve this, we sample subsequent sentences $(\textbf{s}_i, \textbf{s}_{i+1})$ from the training corpus and randomly shuffle their order, as demonstrated in Figure~\ref{fig:self-supervised}c. Then, we train the model to predict the correct order of each pair of shuffled sentences $(\textbf{s}'_i, \textbf{s}'_{i+1})$. The objective of BSOP can be written as
\begin{equation}\small
    \mathcal{L}_{BSOP} = -\frac{1}{N}\sum_{i=1}^{N} \log P(o | \{(\textbf{s}'_i, \textbf{s}'_{i+1})\}_{1}^{2}) ,
\end{equation}
where $o$ represents the correct order between $\textbf{s}'_i$ and $\textbf{s}'_{i+1}$. As illustrated in Figure~\ref{fig:self-supervised}c, we attach a BSOP module to the encoder and follow the same process as in CRL and JP to train the BSOP task.
It is worth noting that BSOP is not a JP subtask, although the two may exhibit the same examples when K=2. In JP, there are always fixed K ({\em e.g.}, 5) sentences. If a context contains more (less) sentences, short (long) ones are merged (split). However, BSOP considers only two consecutive sentences (among all of them) from the contexts. Given this setting, BSOP focuses on modeling direct relations between consecutive sentences, whereas JP focuses on modeling long-term indirect influences scattered across sentences.

\begin{table*}[t!]
\caption{Comparisons between several widely used commonsense QA datasets.}
    \centering
    \begin{tabular}{c|cccccc}
    \hline
        Dataset & Total & Train & Validation & Test & Question Types & Question Type Split \\ \hline
        WIQA & 39,705 & 29,808 & 6,894 & 3,003 & in-paragraph/out-of-paragraph/no-effect & 9,893/17,108/13,694 \\\hline
        CosmosQA &  35,210 & 25,262 & 2,985 & 6,963 & - & - \\\hline
        ReClor & 6,138 & 4,638 & 500 & 1,000 & easy/hard & 440/660   \\ 
    \hline
    \end{tabular}

    \label{tab:dataset}
\end{table*}

\subsubsection{Masked Entity Modeling (MEM)}
To encourage better entity relationship encoding, we propose incorporating an MEM~\cite{zhang-etal-2019-ernie,joshi2020spanbert,guu2020realm} task to learn entity relationships embedded in a context. Let $\textbf{s}$ denote a sequence example, where $\textbf{s}_{\mathcal{E}}$ and $\textbf{s}_{\setminus \mathcal{E}}$ denote the masked entities and remaining unmasked tokens, respectively. The objective of MEM can be written as
\begin{equation}\small
\mathcal{L}_{MEM} = -\frac{1}{N}\sum_{i=1}^{N} \log P(\textbf{s}_{\mathcal{E}} | \textbf{s}_{\setminus \mathcal{E}}).
\end{equation}
As shown in Figure~\ref{fig:self-supervised}d, encoded features of the $\texttt{[CLS]}$ token generated by the encoder are fed to an MEM module to predict entities, and they are further optimized by cross-entropy loss on the masked entity tokens.

\subsubsection{Masked Language Modeling (MLM)}
We also incorporate MLM loss~\cite{lan2019albert} to enhance the model's capability to understand language semantics. This task works well, especially for a scenario where context and questions consist of synonyms, such as ``table'' and ``desk,'' ``improves'' and ``increases,'' ``lead to'' and ``results in.'' Following common practice in MLM~\cite{devlin2018bert}, we randomly choose 15\% of the words in a context sequence and replace them with $\texttt{[MASK]}$. Let $\textbf{s}$ denote a sequence example, where $\textbf{s}_{\mathcal{M}}$ and $\textbf{s}_{\setminus \mathcal{M}}$ denote the masked tokens and the remaining unmasked tokens, respectively. The objective of MLM can be written as
\begin{equation}\small
\mathcal{L}_{MLM} = -\frac{1}{N}\sum_{i=1}^{N} \log P(\textbf{s}_{\mathcal{M}} | \textbf{s}_{\setminus \mathcal{M}}) .
\end{equation}
As shown in Figure~\ref{fig:self-supervised}e, encoded features of the $\texttt{[CLS]}$ token generated by the encoder are fed to an MLM module to predict tokens, then optimized by cross-entropy loss on the masked tokens.

\subsection{Overall Objective}
Finally, the objectives of the QA task and the five presented SSL tasks are combined to optimize the commonsense QA solver. The final objective is written as
\begin{equation}\small
\mathcal{L} = \mathcal{L}_{QA} + \alpha \mathcal{L}_{CRL} + \beta \mathcal{L}_{JP} + \gamma \mathcal{L}_{BSOP} + \lambda \mathcal{L}_{MEM} + \delta \mathcal{L}_{MLM},
\label{eq:overall}
\end{equation}
where $\alpha,\beta,\gamma,\lambda,\delta$ control the importance weight of each SSL objective.

\section{Experiments}
\label{sec: exper}
We evaluate elBERto's effectiveness on commonsense QA datasets and perform an ablation study on individual SSL tasks  and their performance for different question types. Specifically, we examine whether each of the proposed SSL tasks is effective for learning commonsense and whether SSL is able to fuse the five SSL tasks robustly for comprehensive improvement over baseline and individual SSL tasks; this is because an increased number of tasks also increases computational complexity, which may alternatively compromise accuracy.

\subsection{Datasets}

We evaluate the proposed method on three commonsense QA datasets---namely, WIQA~\cite{tandon2019wiqa}, CosmosQA~\cite{huang2019cosmos}, and ReClor~\cite{yu2020reclor}. Table~\ref{tab:dataset} shows the statistics for each dataset. The three datasets can be described as follows:
\begin{itemize}
\item{\verb||} \textbf{WIQA}~\cite{tandon2019wiqa} is a recently proposed dataset asking ``what-if'' questions.
It contains 29,808, 6,894, and 3,003 training, validation, and testing samples, respectively.
Depending on whether the supporting knowledge resides in the given text, the questions are categorized into ``in-paragraph,'' ``out-of-paragraph,'' and ``no-effect'' types. In-paragraph and out-of-paragraph question types refer to examples in which the inquired events or entities can and cannot be found in the context, respectively. No-effect-type questions ask about events or entities not related to the context.
Out-of-paragraph questions are more difficult to solve than in-paragraph ones are because external commonsense knowledge is required.
\item  \textbf{CosmosQA}~\cite{huang2019cosmos} focuses on human daily life scenarios.
It consists of 35,210 samples, which are divided into 25,262, 2,985, and 6,963 training, validation, and testing samples, respectively.
\item{\verb||} \textbf{ReClor}~\cite{yu2020reclor} is collected from standardized graduate admission examinations.
It contains 6,138 samples, which are divided into 4,638, 500, and 1,000 training, validation, and testing samples, respectively.
According to biases contained in the data points, the test set is separated into EASY (Test-E) and HARD (Test-H) sets.
\end{itemize}

%\begin{table}[width=0.83\linewidth,cols=5,pos=h]
\begin{table}
\caption{Test accuracy (\%) on WIQA data. In, out, no represents in-paragraph, out-of-paragraph, no-effect, respectively.}
\centering
\small
\begin{tabular}{@{}lcccc@{}}
\toprule
    Methods & In & Out & No & Total \\
    \cmidrule{1-5}
    Majority ~\citep{tandon2019wiqa} & 45.46 & 49.47 & 0.55 & 30.66 \\
    Polarity~\citep{tandon2019wiqa} & 76.31 & 53.59 & 0.27 & 39.43 \\
    Adaboost~\citep{freund1997-adaboost} & 49.41 & 36.61 & 48.42 & 43.93 \\
Decomp-Att ~\citep{tandon2019wiqa} & 56.31 &  48.56 &  73.42 & 59.48 \\

BERT-base~\citep{devlin2018bert} & 70.57 & 58.54 & 91.08 & 74.26 \\
BERT-large~\citep{devlin2018bert} & 74.91    & 67.08 & 90.20 & 78.12 \\
    RoBERTa-base~\citep{liu2019roberta} &  76.60 & 64.78 & 89.48 & 77.19 \\
    RoBERTa-large~\citep{liu2019roberta} & 76.79 & 66.42 &  91.79 &  78.85  \\

    \red{CURIE (BERT-base)}~\citep{Rajagopal2021CURIEAI} & \red{64.04} & \red{73.58} & \red{90.84} & \red{76.92} \\
    \red{Logic-Guided (RoBERTa-base)} & \red{-} & \red{-} & \red{-} & \red{78.5} \\
\red{REM-Net (RoBERTa-large)}\citep{huang2020rem} & \red{76.23} & \red{69.13} & \red{92.35} & \red{80.09} \\
\red{RGN (RoBERTa-large)}\citep{ijcai2021-0553-RGN} & \red{80.32} & \red{68.63} & \red{91.06} & \red{80.18} \\
\red{PathReasoner (RoBERTa-large)} & \red{77.92} & \red{7.069} & \red{91.55} & \red{80.69} \\

     \cmidrule{1-5}

    \textbf{elBERto (BERT-base)} & 72.83 & 63.86 & \textbf{92.11}& 77.22 \\
    {elBERto (BERT-large)} & 77.74 & 73.48 & 87.73 & 80.19 \\
    {elBERto (RoBERTa-base )} & 78.49 & 74.14 & 86.29 & 79.99  \\
    \textbf{elBERto (RoBERTa-large )}& \textbf{79.06} & \textbf{74.47} &
    88.69 & \textbf{81.22} \\
    
\bottomrule
\end{tabular}
\label{tab:wiqa}
\end{table}

\begin{table}
\caption{Accuracy (\%) on CosmosQA data.}
\centering
\small
\resizebox{1\columnwidth}{!}{
\begin{tabular}{@{}lll@{}}
    \toprule
        Model & Dev & Test  \\
        \cmidrule{1-3}
        Sliding Window~\citep{richardson2013mctest-sliding-window}  & 25.0 & 24.9 \\
        Standford Attentive Reader~\citep{chen2016thorough-standford} & 45.3 & 44.4 \\
        Gated-Attention Reader~\citep{dhingra2016gated} & 46.9 & 46.2 \\
        Co-Matching ~\citep{wang2018co-matching} & 45.9 & 44.7 \\
        Commonsense-Rc~\citep{wang2018yuanfudao-commonsense-RC}  & 47.6 & 48.2 \\
        GPT-FT~\citep{radford2018improving}  & 54.0 & 54.4 \\

        BERT-base~\citep{devlin2018bert} & 60.4 & 62.9 \\
        BERT-large~\citep{devlin2018bert}  & 66.2 & 67.1 \\
        RoBERTo-base ~\citep{liu2019roberta}  & 69.7 & 71.0  \\
        RoBERTa-large ~\citep{liu2019roberta} & 81.2 & 81.7 \\
        ALBERT-xxlarge~\citep{lan2019albert}  & - & 85.4 \\

        KagNet (BERT-base)~\citep{kagnet-emnlp19}  & - & 64.9 \\
        DMCN (BERT-large) ~\citep{zhang2019dual-DMCN} & 67.1 & 67.6 \\
        BERT-large Multiway (BERT-large)~\citep{devlin2018bert}  & 68.3 & 68.4 \\

        \red{SMART (RoBERTa-large)} ~\citep{jiang-etal-2020-smart} & \red{82.0} & \red{81.9} \\
        \red{K-adapter (RoBERTa-large)} ~\citep{wang-etal-2021-k-adapter} & \red{-} & \red{81.8} \\
        \red{GDIN (ALBERT-xxlarge)} ~\citep{tian-etal-2020-scene-gdin} & \red{-} & \red{84.5} \\
        \red{ALICE (ALBERT-xxlarge)} ~\citep{pereira-etal-2020-adversarial} & \red{83.6} & \red{84.6} \\
        \red{RekNet (ALBERT-xxlarge)} ~\citep{zhao2022reference-RekNet} & \red{85.9} & \red{85.7} \\

        \red{REM-Net (RoBERTa-large)} ~\citep{huang2020rem} & \red{-} & \red{81.4} \\

        \cmidrule{1-3}

        {elBERto (BERT-base)} & 65.3 & 66.3 \\
        {elBERto (BERT-large)} &  69.7 & 69.6 \\
        {elBERto (RoBERTo-base)} & 72.0  & 73.6\\
        {elBERto (RoBERTa-large)} & 82.6 & 82.9 \\
        \textbf{elBERto (ALBERT-xxlarge)} & \textbf{86.3} &\textbf{86.1} \\
        \cmidrule{1-3}
        Human & - &  94.0 \\
    \bottomrule
\end{tabular}}
\label{tab:cosmosqa}
\vspace{-5mm}
\end{table}

\begin{table}
\captionof{table}{Accuracy (\%) on the ReClor dataset. Test-E and Test-H represent easy and hard level test questiongs.}
    \centering
    \small
    \begin{tabular}{@{}lcccc@{}}
    \toprule
         Model & Val & Test & Test-E & Test-H  \\
         \cmidrule{1-5}
         FastText ~\citep{bordes2013translating} & 25.0 & 25.0 & 25.0 &25.0 \\
         Bi-LISTM ~\citep{hochreiter1997lstm} & 27.8 & 27.0 & 26.4 & 27.5 \\
         GPT ~\citep{radford2018improving} & 47.6 & 45.4 & 73.0 & 23.8 \\
         GPT-2 ~\citep{radford2019language} & 52.6 & 47.2 & 73.0 & 27.0 \\
         XLNet-base ~\citep{yang2019xlnet} & 55.8 & 50.4 & 75.2 & 30.9 \\
         XLNet-large ~\citep{yang2019xlnet} & 62.0 & 56.0 & 75.7 & 40.5 \\
         BERT-base ~\citep{devlin2018bert} & 54.6 & 47.3 & 71.6 & 28.2 \\
         BERT-large ~\citep{devlin2018bert} & 53.8 & 49.8 & 72.0 & 32.3 \\
         RoBERTa-base ~\citep{liu2019roberta} & 55.0 & 48.5 & 71.1 & 30.7 \\
         RoBERTa-large ~\citep{liu2019roberta} & 62.6 & 55.6 & 75.5 & 40.0 \\

         \cmidrule{1-5}

         elBERto (BERT-base) &  55.4 & 49.1 & 71.4 & 31.6 \\
         elBERto (BERT-large) & 58.4  & 51.8 & 76.6 &  32.3 \\
         \textbf{elBERto (RoBERTa-base)} & 58.4 & 52.0 & \textbf{76.6} & 32.7 \\
         \textbf{elBERto (RoBERTa-large)} & \textbf{65.0} & \textbf{57.3} & 75.7 & \textbf{42.9} \\

         \cmidrule{1-5}
         Human (Graduate Students) & - & 63.0 & 57.1 & 67.2 \\
         \bottomrule
    \end{tabular}
    \label{tab:reclor}
    % \vspace{-3mm}
\end{table}

\begin{table}
    \caption{Comparison of individual self-supervised tasks, elBERto and baseline.}
    \centering
    \huge
    \resizebox{1\columnwidth}{!}{
    \begin{tabular}{@{} c | c c c c c | c c c@{}}
    \toprule
          \# & MLM & MEM & BSOP & JP & CRL & WIQA & CosmosQA & ReClor \\
         \cmidrule{1-9}
          1 &  & & & & & 74.26 & 60.54 & 55.00  \\
          2 & \checkmark & & & & & 75.39 & 62.98 & 55.60 \\
          3 &  & \checkmark & & & & 76.49 & 62.98 & 56.00  \\
          4 &  & & \checkmark & & & 75.66 & 64.42 & 56.80 \\
          5 &  & & & \checkmark & & 76.39 & 63.42 & 57.00 \\
          6 &  & & & & \checkmark & 76.76 & 63.85 & 57.60  \\
          \cmidrule{1-9}
          \red{7} & \red{\checkmark } & \red{\checkmark }&  \red{\checkmark }& & \red{\checkmark }& \red{76.86 }& \red{64.62} & \red{58.00}  \\
          \red{8} & \red{\checkmark }  &  \red{\checkmark }   &  \red{\checkmark } & \red{\checkmark } & & \red{76.66} & \red{64.52} & \red{57.80}  \\
          \red{9} &\red{\checkmark }  &\red{\checkmark }  &\red{\checkmark }  & & & \red{76.59} & \red{64.46} & \red{57.20} \\
          10 &  & & & \checkmark & \checkmark & 77.09 & 64.09 & 58.00  \\
          \cmidrule{1-9}
          11 & \checkmark & \checkmark & \checkmark&
          \checkmark&\checkmark & \textbf{77.22}
          &   \textbf{65.26} & \textbf{58.40}  \\
    \bottomrule
    \end{tabular}}
    \label{tab:ablation1}
\end{table}

\subsection{Baselines}
To verify the effectiveness of elBERTo, we compare it with other methods which utilize the same backbone and training set as ours.
We aim to provide a fair comparison, not to drop some high-performance results on the leaderboard.
However, not all the compared methods use the same backbone, which means they use different backbones to solve CommonsenseQA (one of BERT-base, BERT-large, RoBERTa-base, RoBERTa-large, ALBERT-xxlarge), so we report all 5 elBERTo variants in the paper for comparison including elBERTo(BERT-base), elBERTo(BERT-large), elBERTo(RoBERTa-base), elBERTo(RoBERTa-large) and elBERTo(ALBERT-xxlarge).
Besides, since some methods in the leaderboard use a large amount of corpus to pre-train and obtain high performance, we only compare with those using the same training data.
We introduce all the baselines below based on the dataset. 

On WIQA, we compare with methods mentioned in WIQA~\citep{tandon2019wiqa}, including Majority~\citep{tandon2019wiqa}, Polarity~\citep{tandon2019wiqa}, Adaboost~\citep{freund1995desicion},  Decomp-Att~\citep{tandon2019wiqa}, BERT (base/large) finetune~\citep{devlin2018bert}. Besides, we also compare with RoBERTa (base/large), CURIE~\citep{Rajagopal2021CURIEAI}, Logic-Guided~\citep{asai-hajishirzi-2020-logic}, REM-NET~\citep{huang2020rem}, PathReasoner~\citep{ZHAN-pathreasoner} and RGN~\citep{ijcai2021-0553-RGN}.  Among them, BERT~\citep{devlin2018bert} and RoBERTa~\citep{liu2019roberta} are pre-training models without additional knowledge injection, while REM-NET~\citep{huang2020rem} and PathReasoner~\citep{ZHAN-pathreasoner} utilize external knowledge to facilitate commonsense reasoning. CURIE~\citep{Rajagopal2021CURIEAI} and RGN~\citep{ijcai2021-0553-RGN} utilize graph-based methods for commonsense reasoning and predicting over the commonsense questions. Logic-Guided~\citep{asai2020logic} employs logical and linguistic knowledge to enrich labeled training data and then applies a consistency-based regularizer to train the model.

On CosmosQA, we compare with methods used in CosmosQA~\citep{huang2019cosmos} including  Commonsense-Rc~\citep{wang2018yuanfudao}, GPT-FT~\citep{radford2018improving}, BERT(base/large)~\citep{devlin2018bert}, DMCN~\citep{zhang2019dual}, BERT~\citep{devlin2018bert}, BERT-large Multiway~\citep{huang2019cosmos}, but also compare with recent proposed methods,  as RoBERTa~\citep{liu2019roberta}, ALBERT~\citep{lan2019albert}, KagNet~\citep{lin2019kagnet}, ALICE~\citep{pereira-etal-2020-adversarial}, SMART~\citep{jiang-etal-2020-smart}, K-adapter~\citep{wang-etal-2021-k-adapter}, GDIN~\citep{tian-etal-2020-scene-gdin}, RekNet~\citep{zhao2022reference-RekNet}, REM-NET~\citep{huang2020rem}. Among them, K-adapter~\citep{wang-etal-2021-k-adapter}, GDIN~\citep{tian-etal-2020-scene-gdin}, RekNet~\citep{zhao2022reference-RekNet} and REM-NET~\citep{huang2020rem} use external knowledge to improve model's reasoning ability. KagNet~\citep{lin2019kagnet} utilizes both external knowledge and GNN method. ALICE~\citep{pereira-etal-2020-adversarial} and SMART~\citep{jiang-etal-2020-smart} use data argumentation or regularized optimization methods.

On ReClor, we compare with FastText~\cite{joulin2016bag}, Bi-LISTM~\citep{pennington2014glove}, GPT (1/2)~\cite{radford2018improving}, BERT (base/large)~\citep{devlin2018bert}, XLNet (base/large)~\citep{yang2019xlnet}, RoBERTa (base/large)~\citep{liu2019roberta}.

\subsection{Implementation Details}

\paragraph{Model implementation and optimization.}
We use BERT \cite{devlin2018bert}, RoBERTa \cite{liu2019roberta}, and ALBERT \cite{lan2019albert} as the encoders, with a dropout probability of 0.1. Input sequences are limited to 180 tokens. The learning rates for the BERT-base/RoBERTa-base and BERT-large/RoBERTa-large encoders are 1e-5 and 1e-6, respectively. The warm-up step is set to 10\% of the total training steps.  We train 10 epochs with batch size 4 and Adam optimizer; $\alpha$, $\beta$, $\gamma$, $\lambda$, and $~\delta$ in Eq~\ref{eq:overall} are all equal to $0.2$.

\paragraph{Self-supervised task details.}
For the MLM task, we follow BERT to mask input tokens at random with a probability of 15\%.
If the $i_{th}$ token is chosen, we replace the $i_{th}$ token with [MASK] token 80\% of the time, a random token 10\% of the time and the original $i_{th}$ token 10\% of the time.
For the MEM task, we use TAGME to extract entities because it contains more entities vocabulary compared with the NER tagger, and it has been widely adopted~\cite{asai2019learning,zhao2020transformer-xh,min2020knowledge}. We then mask the entities with 15\% probability.
For the JP task, because samples have different numbers of sentences, we process them such that all samples contain exactly $K$ sentences. For any sample containing sentences not equal to $K$, the following applies: $(\RN{1})$ if the total number of sentences in a sample is smaller than $K$, the longest sentence is decomposed according to punctuation or conjunction words; or $(\RN{2})$ if the total number of sentences is greater than $K$, the two shortest sentences are merged. We repeat $(\RN{1})$ and $(\RN{2})$ until the final number of sentences is $K$, which is set to 5 in our paper.
For BSOP, we randomly select adjacent pairs of sentences and concatenate them with corresponding context sentences individually. Then, we reverse their order with 50\% probability. Sentence pairs in the reversed order are set as negative samples, and sentence pairs in the original order are set as positive samples.
For the CRL task, we develop a list of qualitative words and their corresponding antonyms. \footnote{The antonyms are obtained from \url{https://www.enchantedlearning.com/wordlist/opposites.shtml} and ~\url{https://eslforums.com/antonyms/}, and they contain 658 antonyms. A full list can be downloaded from \url{https://github.com/anonymity2/AntonymsList}. }

\subsection{Experimental Results and Analysis}

\begin{figure}
    \centering
    \includegraphics[trim=0 10 0 0, clip, scale=0.36]{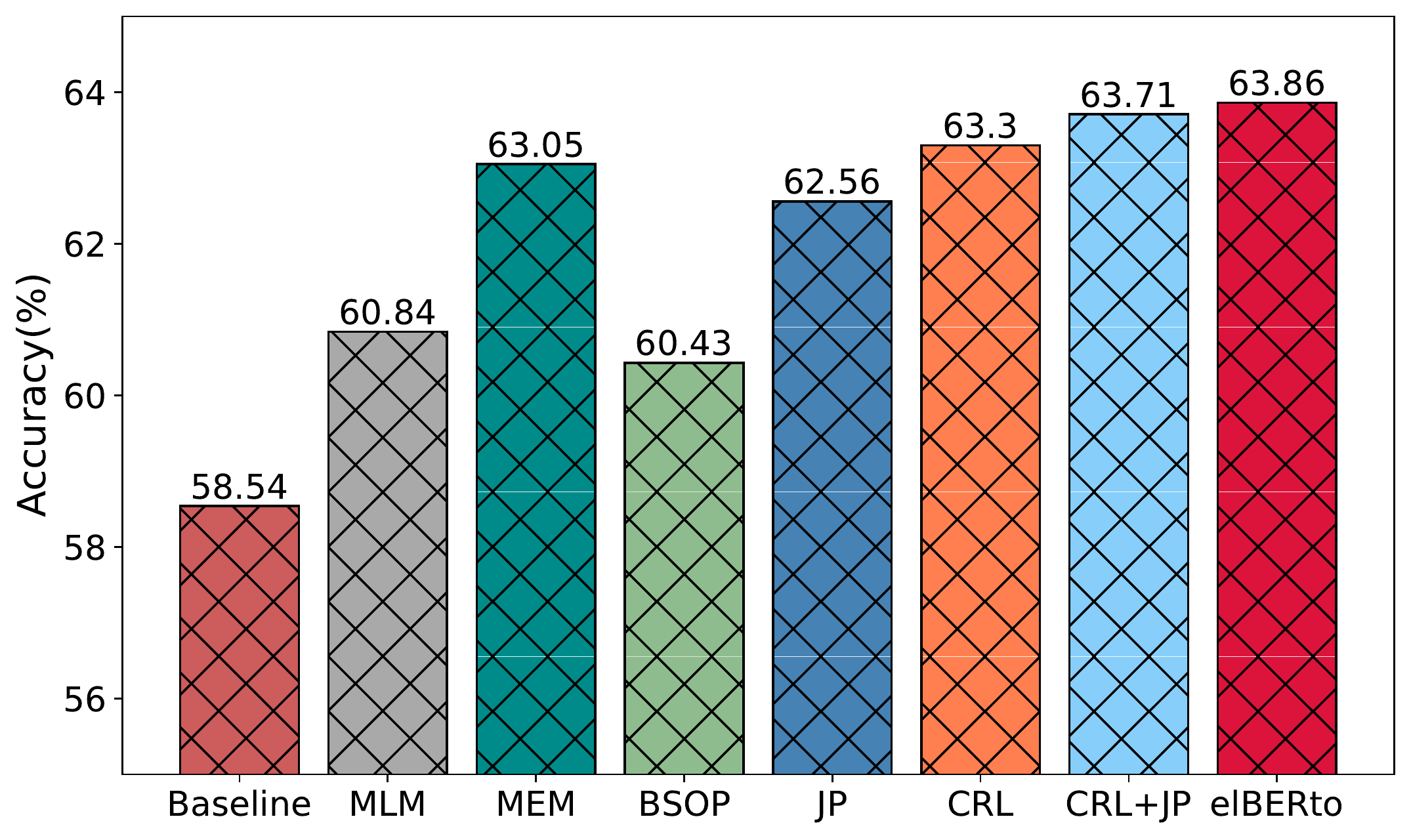}
    \caption{Results on WIQA Out-of-Paragraph questions.}
    \label{fig:ablation2}
\end{figure}

\subsubsection{Comparison with the State-of-the-Art Methods}
First, we verify whether elBERto can effectively exploit additional training signals from existing data and only an external antonym list. As ~\cite{huang2019cosmos} and our experiments show no significant improvement is achieved when further increasing training samples after 15K questions; it is not easy to learn commonsense knowledge by simply stacking more training samples.
Tables~\ref{tab:wiqa},~\ref{tab:cosmosqa}, and~\ref{tab:reclor} demonstrate the experimental results, from which we draw several observations:
\begin{itemize}
    \item On all three datasets, elBERto achieves the new state-of-the-art results compared to all baseline methods that use the same backbone, demonstrating that it successfully learns better commonsense knowledge through limited data. \red{Among all the compared methods, K-adapter~\citep{wang-etal-2021-k-adapter}, RekNet~\citep{zhao2022reference-RekNet}, GDIN~\citep{tian-etal-2020-scene-gdin}, REM-NET \citep{huang2020rem} and PathReasoner~\citep{ZHAN-pathreasoner} utilize additional knowledge to facilitate commonsense reasoning, while CURIE~\citep{Rajagopal2021CURIEAI} and RGN~\citep{ijcai2021-0553-RGN} utilize the graph-based methods. elBERto outperforms these compared approaches by 1.1\%, 0.4\%, 1.6\%, 1.5\%,0.5\% , 0.3\%  and 1.0\% respectively.}
    \item When applied to 12-layer pre-trained models, elBERto achieves a larger improvement than being applied to 24-layer pre-trained models, indicating that it brings more benefits when a less capable encoder is used; this  provides potential advantages for low-resource training.
    % \item \red{Moreover, on both CosmosQA and ReClor leaderboards, elBERto is in first place, exceeding all methods---including those utilizing explicit graph reasoning and external knowledge retrieval, such as KagNet~\citep{lin2019kagnet} (elBERto outperforms it by 1.4\%).}
    \item \red{Moreover, on both CosmosQA and ReClor leaderboards, elBERto exceeds all methods used the same backbones---including those utilizing explicit graph reasoning and external knowledge retrieval. For example, elBERto outperforms KagNet~\citep{lin2019kagnet} by 1.4\%.}
\end{itemize}

\begin{figure*}[!htb]
    \centering
    \includegraphics[trim=0 20 0 0, clip, scale=0.64]{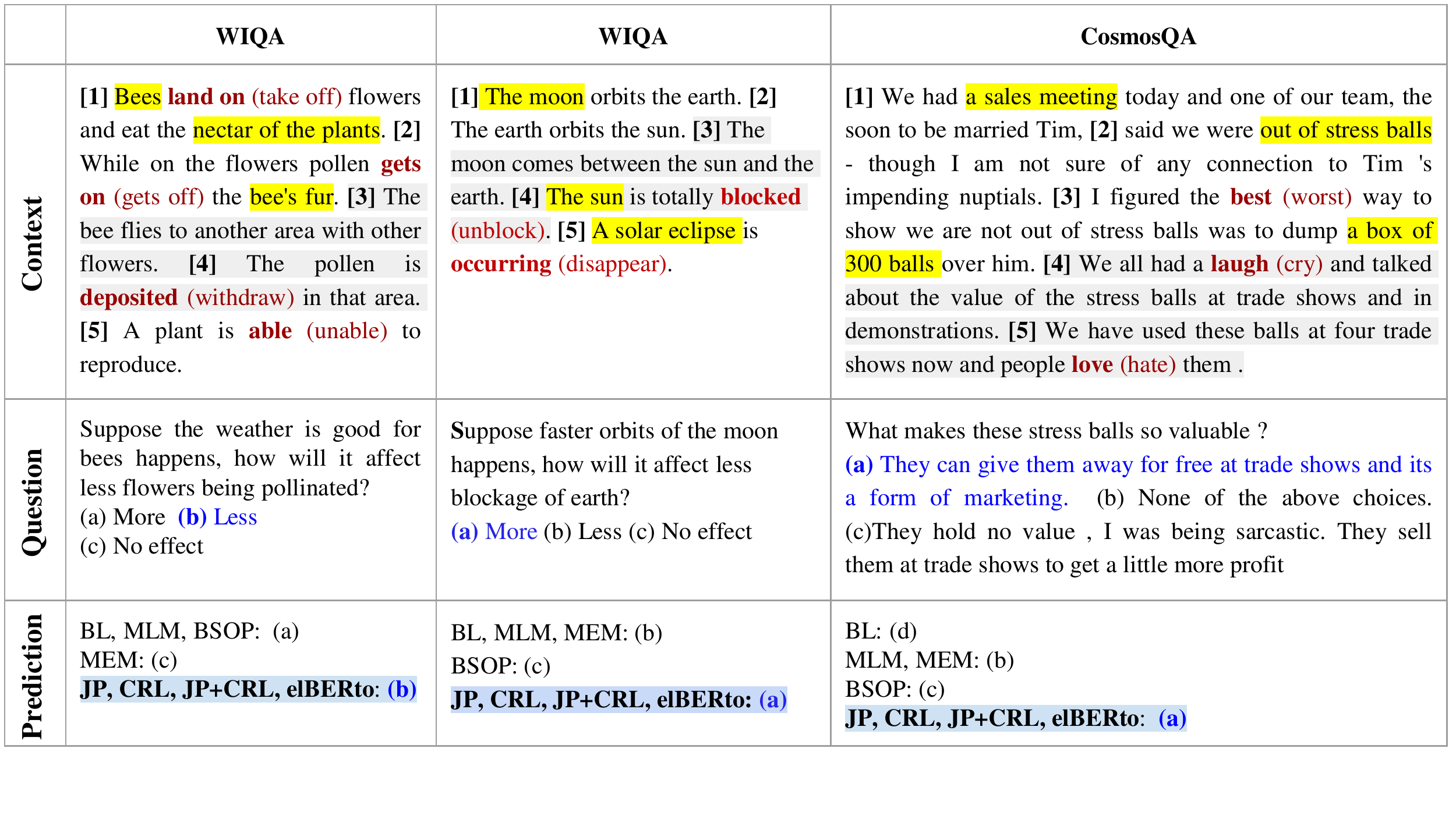}
    \caption{Prediction examples of elBERto, individual SSL tasks, and baseline. Yellow highlighted words are masked entities in MEM. Words in red are used for CRL. Grey-shadowed sentences are used for BSOP. The gold answers are in blue.} % creating contrastive context
    \label{fig:visualize}
    % \vspace{-5mm}
\end{figure*}

\begin{figure*}[!htb]
\center
\includegraphics[angle=0, width=1\textwidth]{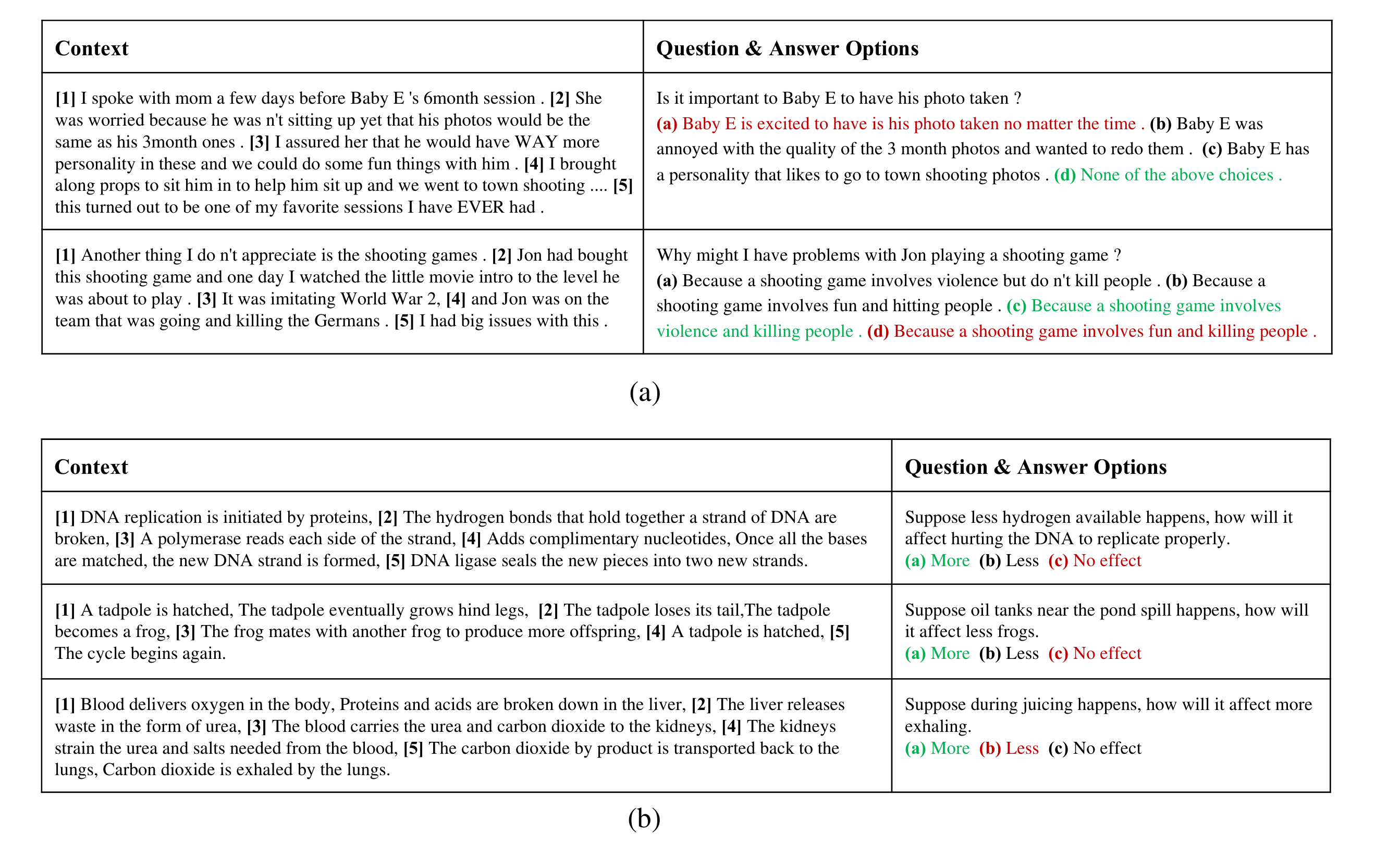}
\caption{Examples of failing cases of elBERto on the WIQA (\textbf{up}), CosmosQA (\textbf{bottom}) datasets. Green text is the ground truth answer. Red text shows the predictions of elBERto. }
 \label{fig:predict_wrong}
\end{figure*}

\subsubsection{Effectiveness Verification of Each Self-Supervised Task}
We study the performance of each proposed task on commonsense QA. The results are demonstrated in Table~\ref{tab:ablation1}, from which we draw several observations:
\begin{itemize}
    \item Each task boosts performance over baseline on all datasets, demonstrating their effectiveness.
    \item The combination of the novel CRL and JP tasks outperforms each individual task on both WIQA and ReClor datasets, and are only 0.33\% lower than BSOP. This shows that joint training of CRL and JP leads to greater benefits for learning commonsense reasoning.
    \item elBERto achieves the best results by combining all five tasks.
\end{itemize}
On the more difficult CosmosQA and ReClor datasets, elBERto leads to the most improvement with 4.82\% and 3.40\% increased accuracy, respectively.
$(\RN{4})$ Even though WIQA might be biased toward the CRL task because its question formation process is similar to CRL's example construction, CRL achieves 3.41\% and 2.60\% accuracy boosts over the best baseline on CosmosQA and ReClor, which have no bias, demonstrating that it indeed effectively learns commonsense knowledge.

\subsubsection{Ablation study}
\red{To further verify the effectiveness of CRL and JP tasks, we show several ablation experimental results in Table 5. Without the CRL task or JP task, the performance of elBERto is dropped, while removing them at the same time would further degrade the performance.  This indicates that the self-supervision tasks of CRL and JP are conducive to improving the model's capabilities to solve Commonsense QA. }

\subsubsection{Evaluation on Out-of-Context and Hard Questions}
We examine elBERto's ability to address hard questions, including those where the answers are not shown in the given context and those containing complex implicit logical rules.
Specifically, on the most difficult out-of-paragraph question type of WIQA, where the answers are out of context and cannot be solved by simple similarity comparison between answers and contexts, elBERto obtains 5.32\% and 6.40\% performance boosts over BERT-base and BERT-large fine-tuning, respectively, as shown in Table~\ref{tab:wiqa} (third column). This verifies that elBERto effectively learns out-of-context commonsense and can use it accurately.
Further, Figure~\ref{fig:ablation2} shows the performance of each task in elBERto, in which CRL, JP, and MEM boost performance over the baseline by 3.41\%, 2.98\% and 2.53\%, respectively.
In addition, on hard-level test samples in the ReClor dataset (Table~\ref{tab:reclor} (last column)), elBERto achieves 2.0\% and 2.9\% accuracy improvements over RoBERTa-base and RoBERTa-large, respectively. Despite using the same pre-trained encoder, elBERto demonstrates superior understanding capability over complex contexts.

\subsubsection{Evaluation on Out-of-Domain Datasets}
In this section, we examine the transferability of elBERto to out-of-domain datasets. Specifically, we train elBERto and BERT on the WIQA dataset comprising scientific procedures and test them on CosmosQA, which contains daily live human conversations, forming distinct source and target domains. elBERto, BERT, and random guessing achieve 31.90\%, 27.87\%, and 25\% accuracy, respectively, demonstrating that elBERto has better transferability to unseen domains.

\subsubsection{Qualitative Results}
Figure~\ref{fig:visualize} demonstrates the visualization results. It can be observed that the baseline fails to predict the correct answer for all examples. JP, CRL, JP+CRL, and elBERto succeed in answering them all correctly, demonstrating their effectiveness for commonsense reasoning in dynamic contexts.

\subsubsection{Failing Case Analysis}
We observe that the performance of the out-of-paragraph question type is lower than the performances for the in-paragraph and no-effect types, although great improvement is achieved in this type. Most failing cases are due to insufficient specialized knowledge, such as knowledge about DNA replication, in the context. Figure~\ref{fig:predict_wrong} demonstrates failing cases of elBERto on the WIQA and CosmosQA datasets.

\section{Conclusions}
 \label{sec: conclusion}
In this chapter, we proposed a novel self-supervised commonsense learning pipeline that better captures commonsense in a complicated multi-hop context and contains the two following novel tasks: $(\RN{1})$ CRL of logical influences between events, achieved by flipping words to their contrastive semantics, and $(\RN{2})$ a new JP task for inferring logical chains in a long context by randomly shuffling sentences. Three classic SSL tasks were also incorporated into the model to improve language encoding ability. On the three challenging commonsense QA datasets, our simple yet effective method achieved the best performance compared with baseline methods, including those utilizing explicit graph reasoning and external knowledge retrieval

\section*{Acknowledgments}

This work is supported by the National Natural Science Foundation of China (Grant No. 61976233 and U19A2073), Guangdong Provincial Natural Science Foundation of China (Grant No. 2019B1515120039), and the Office of Naval Research under grant N00014-18-1-2871.

{\small
\bibliographystyle{ieee_fullname}
\bibliography{egbib}
}

\end{document}